\documentclass[runningheads]{llncs}

\usepackage{silence}
\usepackage[T1]{fontenc}
\usepackage{graphicx}
\usepackage{amsmath}
\usepackage{multirow}
\usepackage{placeins}
\usepackage{amssymb}
\usepackage{bm}
\usepackage{booktabs}
\usepackage{url}

\usepackage{tikz}
\usepackage{marvosym}
\usetikzlibrary{positioning, arrows.meta}
\begin{document}

\title{Optimal Multi-way Decision Trees for Stratified Sampling in Online Controlled Experiments}
\titlerunning{Optimal Multi-way Decision Trees for Stratified Sampling}

\author{Tomoka Takei\inst{1}\textsuperscript{\Letter}\orcidID{0009-0008-6623-0585} \and
Shunnosuke Ikeda\inst{1}\orcidID{0009-0004-4283-0819} \and
Yuichi Takano\inst{1}\orcidID{0000-0002-8919-1282}}

\authorrunning{T. Takei et al.}

\institute{University of Tsukuba, Tsukuba-shi, Ibaraki 305-8573, Japan\\
\email{\{s2620461@u, ikeda@cs, ytakano@sk\}.tsukuba.ac.jp}}

\maketitle

\begin{abstract}
Online controlled experiments, or A/B tests, are widely used to estimate causal effects on digital platforms. A central challenge is to improve experimental sensitivity, or statistical power, without increasing the experimental sample size. Stratified sampling is a classical variance reduction technique; however, its effectiveness depends critically on how the strata are constructed.
We thus propose an optimization-based stratification framework for stratified sampling using optimal multi-way decision trees. Our method, called Optimal Multi-way Stratification Trees (OMST), formulates stratification as a path-selection problem over a feature graph. The selected paths define interpretable stratification rules and are optimized using an exact variance-minimizing binary optimization formulation under continuous proportional allocation and a Neyman-type optimal allocation. We incorporate supervised optimal binning to generate outcome-relevant candidate splits for numerical features. Furthermore, we introduce reduction procedures for redundant candidate paths and assignment constraints, substantially reducing the optimization problem size. 
Experiments on both a real-world and a simulated dataset demonstrate that OMST achieves comparable or superior variance reduction to existing methods while maintaining shallow and interpretable stratification trees.
\end{abstract}

\section{Introduction}

\subsection{Background}
Online controlled experiments (OCEs), also known as A/B tests, are widely used on digital platforms to estimate the causal effects of treatments such as marketing campaigns, item recommendations, and user interface changes. By randomly assigning users to treatment and control groups, OCEs enable unbiased estimation of treatment effects and provide a reliable basis for data-driven decision making.
A fundamental challenge in OCEs is to improve the sensitivity (or statistical power) of experiments. In large-scale web services, even small improvements in key metrics can have a significant business impact, which necessitates conducting experiments with high statistical sensitivity~\cite{xie2016improving}. Since treatment and control samples are typically independent, enhancing the sensitivity relies heavily on reducing the variance of the estimator. Although increasing the sample size is the most straightforward way to reduce this variance, it is often impractical due to cost, time constraints, and potential negative impacts on user experience. Therefore, variance reduction techniques have been extensively studied to enhance estimation accuracy without increasing the sample size.

\subsection{Related Work}
Existing variance reduction methods for controlled experiments can be broadly categorized into regression-based and sampling-based approaches. Regression-based methods improve the estimator after assignment to the treatment and control groups, whereas sampling-based methods construct a more efficient experimental design before or during assignment.

Regression-based approaches, such as CUPED (Controlled Experiments Using Pre-Experiment Data), reduce variance by adjusting outcome variables using covariates through linear regression~\cite{deng2013improving}. Recent extensions incorporate machine learning models to improve predictive accuracy~\cite{guo2021machine} and robust estimators such as trimmed means~\cite{charette2025improving}. While effective when strong predictive covariates are available, these approaches rely on model assumptions and may suffer from model misspecification~\cite{freedman2008regression}.

Sampling-based approaches include stratified sampling, which partitions the population into strata (i.e., homogeneous groups) and allocates samples within each group to reduce within-stratum variation~\cite{Cochran1977,friedrich2015fast}. Related designs such as post-stratification also exploit strata after sampling to improve estimation~\cite{holt1979post}. In practice, strata are often constructed using heuristic methods such as $K$-means clustering~\cite{kim2013stratified,steinley2006k,Tipton2014}. However, clustering-based methods minimize the intra-cluster variance in the covariate space rather than the variance of the stratified estimator itself, which can be suboptimal for variance reduction. Momozu et al.~\cite{momozu2025subset} selected stratification variables based on variance reduction criteria and then applied clustering. However, such two-stage approaches do not directly optimize the stratification structure.

To improve the interpretability of stratification rules, decision trees such as CART~\cite{breiman1984cart} have been used to approximate clustering-based strata~\cite{kim2013stratified}. Stratification trees were also proposed to jointly determine stratification and treatment assignment~\cite{tabord2023stratification}. However, these approaches rely on heuristic or local optimization, and stratification trees additionally require pilot data. Moreover, conventional binary trees can create deeply nested rules, reducing their interpretability.

Recently, optimization-based approaches for learning decision trees have been proposed~\cite{costa2023recent}, including optimal classification trees~\cite{bertsimas2017optimal} and flow-based formulations~\cite{aghai2021flowoct}. In particular, Subramanian and Sun~\cite{subramanian2023scalable} introduced Optimal Multiway-split Decision Trees (OMT), which are formulated as binary optimization problems. Unlike binary trees, multi-way decision trees represent multiple branches at a node, allowing shallower and more concise partitioning rules. By framing tree construction as path selection on a feature graph, OMT explicitly represents rules and optimizes tree structures under combinatorial constraints.
However, these optimization-based tree learning methods have been developed primarily for machine learning tasks~\cite{subramanian2023scalable,suzuki2026interpretable} and have not been extended to optimize stratified sampling designs for variance reduction.

\subsection{Our Contribution}
We propose an optimization-based framework that directly links tree-structured stratification with the variance formula of stratified sampling. 
First, we formulate stratification as a path-selection problem for Optimal Multi-way Stratification Trees (OMST), in which the selected paths define interpretable strata.
Second, we integrate classical sample allocation theory into our optimization framework to determine the sample size drawn from each stratum. Specifically, we incorporate these allocation strategies into the optimization framework and derive binary optimization formulations tailored to two representative allocation schemes: proportional allocation (where sample sizes are determined relative to stratum sizes) and Neyman-type optimal allocation (where sample sizes are optimized for variance reduction).
Third, we incorporate supervised optimal binning~\cite{navas2022optimal} to generate informative candidate splits for numerical features. 
Fourth, we introduce reduction procedures for redundant candidate paths and equivalent assignment patterns to improve computational scalability.
Together, these contributions position OMST as a principled alternative to heuristic clustering and greedy tree construction for variance reduction in online controlled experiments.

To validate the effectiveness of our method, we conducted computational experiments using two datasets with different covariate structures. The experimental results show that our OMST outperformed or matched existing methods, including $K$-means clustering, CART, and CUPED. In particular, our method delivered strong variance reduction with shallow multi-way decision trees, demonstrating that interpretability can be preserved while achieving substantial improvements in estimation accuracy relative to simple random sampling. The results also show that supervised optimal binning improved stratification quality relative to standard quantile-based binning and that our reduction procedures decreased the number of decision variables and constraints substantially.

\section{Online Controlled Experiments}
\label{sec:online_experiments}

On digital platforms, enhancing the sensitivity (or statistical power) of controlled experiments is essential for detecting subtle yet economically significant treatment effects. Let $\bar{Y}^{(t)}$ and $\bar{Y}^{(c)}$ be the sample means of the outcome variable for the treatment and control groups, respectively. The average treatment effect (ATE) is estimated by the difference in sample means:
\begin{equation}
\label{eq:ate_estimator}
\hat{\tau} := \bar{Y}^{(t)}-\bar{Y}^{(c)}.
\end{equation}

Assuming that the treatment and control assignments are statistically independent, the variance of the estimator decomposes as:
\begin{equation}
\label{eq:variance_decomposition}
\mathrm{Var}(\hat{\tau}) = \mathrm{Var}(\bar{Y}^{(t)})+\mathrm{Var}(\bar{Y}^{(c)}),
\end{equation}
which shows that improving sensitivity requires reducing outcome variance within each group.
Accordingly, we develop the method for accurately estimating one population mean; applying the same stratified design within each group reduces the two variance components in Eq.~\eqref{eq:variance_decomposition} and therefore the variance of the ATE estimator.

Let $[m] := \{1, 2, \ldots, m\}$ be the set of consecutive positive integers for any $m \in \mathbb{Z}_+$. To analyze this variance formally, we adopt a finite-population sampling perspective~\cite{Cochran1977}. Consider a target population of size $N$, where each unit $i \in [N]$ has an associated outcome $Y_i$.
Under simple random sampling without replacement (SRS), the sample mean $\bar{Y}$ based on a sample of size $n$ has the variance:
\begin{equation}
\label{eq:srs_variance}
\mathrm{Var}_{\mathrm{SRS}}(\bar{Y}) = \left(1-\frac{n}{N}\right)\frac{S^2}{n},
\end{equation}
where $S^2 := (N-1)^{-1}\sum_{i\in[N]}(Y_i-\mu)^2$ denotes the finite-population variance, and $\mu := N^{-1}\sum_{i\in[N]}Y_i$ is the population mean. Under standard random sampling, reducing variance relies on increasing $n$, which is often costly, slow, or impractical.

\section{Stratified Sampling}
\label{sec:stratified_sampling}

This section describes the three components of stratified sampling: stratification, calculation of the sample mean, and sample allocation.

\subsection{Stratification}

Stratification divides the population into $H$ disjoint strata, or homogeneous subgroups, based on stratification variables (e.g., covariates such as age or gender). Each stratum $h\in[H]$ consists of $N_h$ units, satisfying $N = \sum_{h\in[H]} N_h$. Let $\mu_h$ and $S_h^2 := (N_h-1)^{-1}\sum_{i\in[N_h]}(Y_i-\mu_h)^2$ denote the population mean and the finite-population variance within stratum $h\in[H]$, respectively. To achieve effective variance reduction, the constructed strata should be as homogeneous as possible with respect to the outcome.

\subsection{Calculation of the Sample Mean}

Let $n_h$ be the sample size allocated to stratum $h \in [H]$, and $\bar{Y}_h$ be the corresponding sample mean obtained within that stratum. The stratified estimator of the overall population mean is defined as the population-weighted average of the stratum-specific sample means:
\begin{equation}
\label{eq:strat_estimator}
\hat{\mu}_{\mathrm{strat}} = \sum_{h\in[H]}\frac{N_h}{N}\bar{Y}_h.
\end{equation}
Under independent random sampling without replacement within each stratum, the variance of $\hat{\mu}_{\mathrm{strat}}$, incorporating the finite population correction, is given by Cochran~\cite{Cochran1977} as follows:
\begin{equation}
\label{eq:var_strat}
\mathrm{Var}(\hat{\mu}_{\mathrm{strat}}) = \frac{1}{N^2} \sum_{h\in[H]} \frac{N_h^2 S_h^2}{n_h} - \frac{1}{N^2} \sum_{h\in[H]} N_h S_h^2.
\end{equation}
The first term of Eq.~\eqref{eq:var_strat} depends directly on the chosen sample sizes $n_h$, whereas the second term is independent of the sample allocation and is determined solely by the stratum configuration and the overall population size.

\subsection{Sample Allocation}
\label{sec:Sample_Allocation}

Sample allocation determines the specific sample size $n_h$ to be drawn from each stratum $h\in[H]$ under the total sample size constraint $\sum_{h\in[H]} n_h = n$. Under proportional allocation, samples are assigned in proportion to the size of each stratum:
\begin{equation}
\label{eq:prop_alloc}
n_h = \frac{N_h}{N}n \qquad (h\in[H]).
\end{equation}
While simple and robust, proportional allocation ignores heterogeneity in the within-stratum variances $S_h^2$.

Optimal allocation instead determines the allocation vector $\bm{n} := (n_h)_{h \in [H]} \in \mathbb{Z}_+^H$ so as to minimize the variance in Eq.~\eqref{eq:var_strat} for a given total sample size $n$. This can be formulated as an integer optimization problem, as discussed by Friedrich et al.~\cite{friedrich2015fast}:
\begin{align}
\min_{\bm{n}\in\mathbb{Z}_+^H}\quad
& \sum_{h\in[H]}\frac{N_h^2 S_h^2}{n_h} \label{eq:allocation_obj}\\
\mathrm{s.~t.}\quad
& \sum_{h\in[H]} n_h = n,\\
& l_h \le n_h \le u_h \qquad (h\in[H]), \label{eq:allocation_st}
\end{align}
where $l_h$ and $u_h$ denote the pre-specified lower and upper bounds for the sample size in stratum $h\in[H]$ (typically $l_h \ge 1$ and $u_h = N_h$). If the integrality and boundary constraints are relaxed, the analytical solution yields the classical Neyman allocation~\cite{Neyman1934}:
\begin{equation}
\label{eq:neyman}
n_h = \frac{N_h S_h}{\sum_{k\in[H]}N_k S_k}n \qquad (h\in[H]),
\end{equation}
which assigns more samples to strata that are larger or exhibit greater variability.

\section{Optimal Stratification}

This section presents our optimization framework for stratified sampling. The key idea is to generate interpretable candidate strata as directed paths on a feature graph and then select a subset of paths that minimizes the variance of the stratified estimator.

\subsection{Optimal Multi-way Stratification Trees}

Our method, {\em Optimal Multi-way Stratification Trees} (OMST), builds on Optimal Multiway-split Decision Trees (OMT)~\cite{subramanian2023scalable}, which employ a directed acyclic \textit{feature graph}. In this graph, each feature forms a layer, and each node corresponds to a feature value or interval. Adjacent layers are fully connected.

\begin{figure}[tb]
\centering
\begin{minipage}{0.49\linewidth}
\centering
\resizebox{\linewidth}{!}{%
\begin{tikzpicture}[
    node distance=1.0cm and 2.4cm,
    mynode/.style={circle, draw, minimum size=1.3cm, inner sep=1pt, align=center, font=\footnotesize},
    rootnode/.style={mynode, fill=gray!5},
    gendernode/.style={mynode, fill=orange!15},
    incomenode/.style={mynode, fill=blue!10},
    freqnode/.style={mynode, fill=green!10},
    skipnode/.style={mynode, fill=gray!5},
    thickarrow/.style={-Stealth, line width=1.8pt},
    thinarrow/.style={-Stealth, thin, opacity=0.75}
]

\node[rootnode] (source) {Source};

\node[gendernode, above right=0.65cm and 1.2cm of source] (gen_m) {Male};
\node[gendernode, below=0.25cm of gen_m] (gen_f) {Female};
\node[skipnode, below=0.25cm of gen_f] (s_gen) {Skip};

\node[incomenode, above right=1.4cm and 1.7cm of gen_f] (inc_h) {High};
\node[incomenode, below=0.25cm of inc_h] (inc_m) {Medium};
\node[incomenode, below=0.25cm of inc_m] (inc_l) {Low};
\node[skipnode, below=0.25cm of inc_l] (s_inc) {Skip};

\node[freqnode, above right=1.3cm and 1.7cm of inc_m] (freq_vh) {Very\\high};
\node[freqnode, below=0.18cm of freq_vh] (freq_h) {High};
\node[freqnode, below=0.18cm of freq_h] (freq_m) {Medium};
\node[freqnode, below=0.18cm of freq_m] (freq_l) {Low};
\node[skipnode, below=0.18cm of freq_l] (s_freq) {Skip};

\node[rootnode, below right=2.05cm and 1.2cm of freq_vh] (sink) {Sink};

\foreach \n in {gen_m, gen_f, s_gen} \draw[thinarrow] (source) -- (\n);
\foreach \u in {gen_m, gen_f, s_gen} {
    \foreach \v in {inc_h, inc_m, inc_l, s_inc} \draw[thinarrow] (\u) -- (\v);
}
\foreach \u in {inc_h, inc_m, inc_l, s_inc} {
    \foreach \v in {freq_vh, freq_h, freq_m, freq_l, s_freq} \draw[thinarrow] (\u) -- (\v);
}
\foreach \n in {freq_vh, freq_h, freq_m, freq_l, s_freq} \draw[thinarrow] (\n) -- (sink);

\draw[thickarrow] (source) -- (gen_m);
\draw[thickarrow] (source) -- (gen_f);
\draw[thickarrow] (gen_m) -- (inc_h);
\draw[thickarrow] (gen_m) -- (inc_m);
\draw[thickarrow] (gen_f) -- (inc_l);
\draw[thickarrow] (inc_l) -- (freq_vh);
\draw[thickarrow] (inc_l) -- (freq_h);
\draw[thickarrow] (inc_l) -- (freq_m);
\draw[thickarrow] (inc_h) -- (s_freq);
\draw[thickarrow] (inc_m) -- (s_freq);
\draw[thickarrow] (freq_vh) -- (sink);
\draw[thickarrow] (freq_h) -- (sink);
\draw[thickarrow] (freq_m) -- (sink);
\draw[thickarrow] (s_freq) -- (sink);

\node[below=1.46cm of s_gen, font=\bfseries\normalsize] {Gender};
\node[below=0.66cm of s_inc, font=\bfseries\normalsize] {Income};
\node[below=0.05cm of s_freq, font=\bfseries\normalsize] {Purchase frequency};

\end{tikzpicture}%
}

\vspace{1mm}
{\footnotesize (a) Feature graph}
\end{minipage}
\hfill
\begin{minipage}{0.46\linewidth}
\centering
\resizebox{\linewidth}{!}{%
\begin{tikzpicture}[
    every node/.style={draw, circle, minimum size=1.3cm, inner sep=1pt, align=center, font=\footnotesize},
    edge from parent/.style={draw, -{Stealth}},
    level 1/.style={sibling distance=2.5cm, level distance=1.8cm},
    level 2/.style={sibling distance=1.6cm, level distance=1.8cm},
    level 3/.style={sibling distance=1.6cm, level distance=1.8cm},
    rootnode/.style={circle, draw, fill=gray!5, minimum size=1.3cm, inner sep=1pt, align=center, font=\footnotesize},
    gendernode/.style={rectangle, draw, fill=orange!15, minimum size=1.1cm, inner sep=1pt, align=center, font=\footnotesize},
    incomenode/.style={rectangle, draw, fill=blue!10, minimum size=1.1cm, inner sep=1pt, align=center, font=\footnotesize},
    freqnode/.style={circle, draw, fill=green!10, minimum size=1.3cm, inner sep=1pt, align=center, font=\footnotesize},
    label_style/.style={draw=none, fill=none, font=\bfseries\small, anchor=west}
]

\node[rootnode] (root) {Source}
    child { node(s_gen) [gendernode] {Male}
        child { node[incomenode] {High} }
        child { node[incomenode] {Medium} }
    }
    child { node[gendernode] {Female}
        child { node(s_inc) [incomenode] {Low}
            child { node(s_freq) [freqnode] {Very\\high} }
            child { node[freqnode] {High} }
            child { node[freqnode] {Medium} }
        }
    };

\coordinate (right_edge) at (3.66,0);
\node[label_style] at (right_edge |- s_gen)  {Gender};
\node[label_style] at (right_edge |- s_inc)  {Income};
\node[label_style] at (right_edge |- s_freq) {Purchase frequency};

\end{tikzpicture}%
}

\vspace{-1mm}
{\footnotesize (b) Selected multi-way decision tree}
\end{minipage}

\caption{Candidate path selection in OMST. Thick arrows indicate selected paths, and each leaf corresponds to a stratum.}
\label{fig:omt_example}
\end{figure}
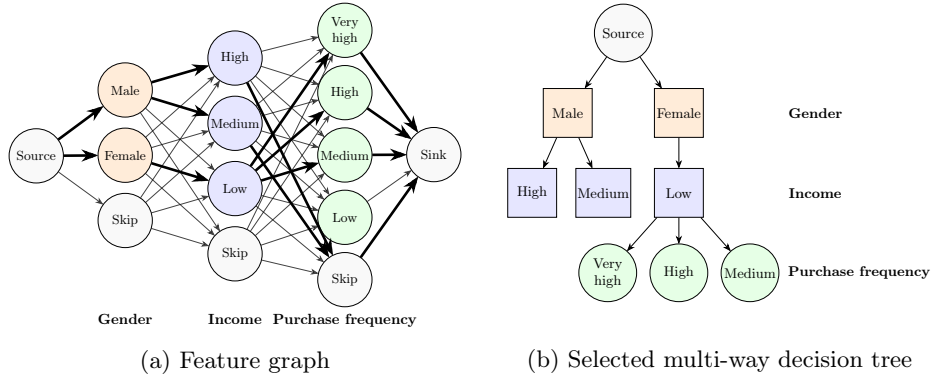

Figure~\ref{fig:omt_example}(a) illustrates a feature graph with Gender, Income, and Purchase frequency. Categories or predefined intervals become nodes. Each directed path from {\em Source} to {\em Sink} is a conjunction of split conditions (e.g., Gender = Male and Income = High), defining one candidate stratum. \textit{Skip} nodes indicate that the corresponding features are not used in split conditions.

A multi-way decision tree is constructed by selecting paths from {\em Source} to {\em Sink}. The thick arrows in Fig.~\ref{fig:omt_example}(a) yield the tree in Fig.~\ref{fig:omt_example}(b). Each leaf is a stratum defined by path conditions, providing an interpretable stratification rule for stratified sampling.

\subsection{Optimal Binning for Numerical Features}

Our framework depends on candidate split quality. For numerical features, a standard baseline is quantile binning~\cite{dougherty1995supervised}, which partitions data into bins with approximately equal sample sizes. This unsupervised approach is stable but may miss outcome-relevant thresholds.

We therefore adopt supervised optimal binning~\cite{navas2022optimal}, which incorporates outcome information. First, a fine set of pre-bins is constructed, for example, using a decision tree-based pre-binning procedure. Then, adjacent pre-bins are merged to form final bins so as to maximize a binning quality score while satisfying practical constraints such as minimum bin size and, if necessary, monotonicity or statistical significance constraints.

A desirable binning separates bin means while controlling within-bin variability, aligning with variance reduction. Thus, optimal binning improves the quality of the feasible splits by generating informative candidate conditions.
In our method, we use features with high binning quality scores to construct candidate paths, thereby reducing the number of candidate paths while focusing on outcome-relevant features.

For a discretized numerical feature partitioned into $B$ bins with cut points $b_1,b_2,\ldots,b_{B-1}$, candidate intervals are generated as
\begin{equation}
\left\{(b_s,b_t] \;\middle|\; s,t\in\{0,1,\ldots,B\},\ s<t,\ (s,t)\neq(0,B)\right\},
\end{equation}
where $b_0=-\infty$ and $b_B=\infty$.

For categorical variables, grouped categories and their complements are candidate conditions, allowing both fine-grained and coarse stratification rules.

\subsection{Variance Minimization Formulation}

The stratification design problem is formulated as a combinatorial optimization problem that uses a variance-derived objective to select an optimal subset of directed paths on a feature graph. 

Let $\mathcal{P}$ denote the set of candidate paths, each corresponding to a potential stratum. For path $j\in\mathcal{P}$, let $z_j\in\{0,1\}$ indicate whether path $j$ is selected, let $N_j$ and $S_j^2$ be the population size and within-path variance, and let $a_{ij}\in\{0,1\}$ indicate whether unit $i\in[N]$ belongs to path $j$. The OMST formulation imposes the following common constraints:
\begin{align}
& \sum_{j\in\mathcal{P}} a_{ij} z_j \le 1
\qquad (i\in[N]), \label{eq:assign}\\
& \sum_{j\in\mathcal{P}} N_j z_j \ge \rho N, \label{eq:coverage}\\
& \sum_{j\in\mathcal{P}} z_j \le L, \label{eq:leaf}\\
& z_j \in \{0,1\}
\qquad (j\in\mathcal{P}), \label{eq:binary}
\end{align}
where $L$ denotes the pre-specified maximum number of selected strata. Constraint \eqref{eq:assign} ensures that each unit belongs to at most one selected stratum. Constraint \eqref{eq:coverage} requires at least a fraction $\rho \in (0, 1]$ of the population to be covered by selected paths; in particular, when $\rho=1$, all units must be covered by the selected paths. Finally, constraint \eqref{eq:leaf} limits the total number of selected strata.

If path $j \in \mathcal{P}$ is selected, the corresponding stratum contributes
\begin{equation}
\label{eq:stratum_contribution}
\left(\frac{N_j}{N}\right)^2
\left(1-\frac{n_j}{N_j}\right)
\frac{S_j^2}{n_j}
z_j
\end{equation}
to the variance of the stratified sample mean (see Eq.~\eqref{eq:var_strat}), where $n_j$ is the allocated sample size. The overall design problem depends on the sample allocation methods described in Sec.~\ref{sec:Sample_Allocation}.

\subsubsection{Proportional Allocation}

Under proportional allocation, the sample size of each stratum is given by Eq.~\eqref{eq:prop_alloc}. Substituting this expression into Eq.~\eqref{eq:stratum_contribution} and summing over all candidate paths $j \in \mathcal{P}$, we obtain
\begin{equation}
\mathrm{Var}(\hat{\mu}_{\mathrm{strat}})
=
\left(1-\frac{n}{N}\right)\frac{1}{nN}
\sum_{j\in\mathcal{P}} N_j S_j^2 z_j.
\end{equation}
Since the multiplicative coefficient is constant, minimizing the variance reduces to the following binary linear optimization problem:
\begin{equation}
\label{eq:mip_prop}
\begin{aligned}
\min_{\bm{z}}\quad
& \sum_{j\in\mathcal{P}} N_j S_j^2 z_j\\
\mathrm{s.~t.}\quad
& \mathrm{Eqs.}~\eqref{eq:assign}\text{--}\eqref{eq:binary}.
\end{aligned}
\end{equation}

Thus, proportional allocation selects strata minimizing aggregate within-stratum variance weighted by population size.
Eq.~\eqref{eq:mip_prop} gives the exact variance objective when $\rho=1$ and the continuous proportional sample sizes $n_j=nN_j/N$ are feasible without rounding.

\subsubsection{Optimal Allocation}

When sample allocation is also optimized, jointly determining path selection and integer sample sizes leads to a nonlinear integer optimization problem because the variance contains terms involving $1/n_j$.
To maintain tractability, we adopt the Neyman allocation given by Eq.~\eqref{eq:neyman} as a surrogate for evaluating candidate strata. Specifically, substituting the Neyman allocation into Eq.~\eqref{eq:stratum_contribution} and summing over all candidate paths $j \in \mathcal{P}$ yields
\begin{equation}
\mathrm{Var}(\hat{\mu}_{\mathrm{strat}})
=
\frac{1}{N^2}
\left[
\frac{1}{n}
\left(\sum_{j\in\mathcal{P}}N_j S_j z_j\right)^2
-
\sum_{j\in\mathcal{P}}N_j S_j^2 z_j
\right].
\end{equation}
Ignoring the constant factor $1/N^2$, the path-selection problem is formulated as the following binary quadratic optimization problem:
\begin{equation}
\label{eq:mip_opt}
\begin{aligned}
\min_{\bm{z}}\quad
& \frac{1}{n}\left(\sum_{j\in\mathcal{P}}N_j S_j z_j\right)^2
- \sum_{j\in\mathcal{P}}N_j S_j^2 z_j\\
\mathrm{s.~t.}\quad
& \mathrm{Eqs.}~\eqref{eq:assign}\text{--}\eqref{eq:binary}.
\end{aligned}
\end{equation}
After selecting strata, the actual integer allocation is computed by solving the integer optimization problem~\eqref{eq:allocation_obj}\text{--}\eqref{eq:allocation_st}.
Eq.~\eqref{eq:mip_opt} gives the exact variance objective when every population unit belongs to exactly one selected path and the continuous Neyman allocation is feasible.

\subsection{Problem-Size Reduction}
\label{sec:reduction}
The main computational challenge in the path-based formulation is the number of candidate paths. 
Let $p$ be the total number of features, with each feature represented by at most $B$ nodes. 
If the tree depth is bounded by $d$, the number of candidate paths $|\mathcal{P}|$ is bounded as follows~\cite{subramanian2023scalable}:
\begin{equation}
\label{eq:path_count}
|\mathcal{P}| = \mathcal{O}\left( \binom{p}{d} B^d \right).
\end{equation}

Eq.~\eqref{eq:path_count} indicates combinatorial growth in $p$, $d$, and $B$, whereas $N$ mainly increases the size of the assignment system.
Because Gurobi uses a branch-and-bound-type procedure, runtime is instance dependent and may grow exponentially with the number of binary variables in the worst case. The proposed reductions are therefore particularly important for high-dimensional instances.

To mitigate this exponential growth, we apply three preprocessing procedures to reduce the problem size before optimization.

\begin{description}
  \item[Path Pruning via Stratum Size:] 
  Let $N_j$ be the number of units in the stratum corresponding to candidate path $j \in \mathcal{P}$. Because $S_j^2$ is unstable for tiny strata, we prune every path satisfying $N_j<N_{\min}$.
  Increasing $N_{\min}$ removes more binary variables and stabilizes variance estimates, but may discard small strata that provide useful variance reduction; decreasing it has the opposite computational--statistical trade-off.

  \item[Variable Reduction via Path Aggregation:] 
  Distinct split sequences can yield identical unit subsets. For each path $j\in\mathcal{P}$, let $\bm{a}_{\cdot j}:=(a_{ij})_{i\in[N]}$ be its inclusion vector. We group paths with $\bm{a}_{\cdot j}=\bm{a}_{\cdot j'}$ and retain from each group the representative with the fewest split conditions.

  \item[Constraint Reduction via Unit Aggregation:] 
  Let $\bm{a}_{i\cdot}:=(a_{ij})_{j\in\mathcal{P}}$ be the assignment vector for unit $i \in [N]$. We aggregate units with identical vectors (i.e., $\bm{a}_{i\cdot} = \bm{a}_{i'\cdot}$) into $G$ disjoint groups $\mathcal{G}_g$ for $g\in[G]$, where $G\ll N$. Let $\bm{a}^{(g)}:= (a^{(g)}_j)_{j \in \mathcal{P}}$ denote the representative pattern of group $\mathcal{G}_g$. The original $N$ constraints in Eq.~\eqref{eq:assign} are then compressed into $G$ constraints:
  \begin{equation}
  \sum_{j\in\mathcal{P}} a^{(g)}_{j} z_j \le 1 \qquad (g\in[G]).
  \end{equation}
  This reduction preserves the exact mathematical structure of the original problem while drastically reducing its size.
\end{description}

Let $\widetilde{\mathcal{P}}$ denote the set of candidate paths remaining after path pruning and path aggregation. Before reduction, the optimization problems~\eqref{eq:mip_prop} and~\eqref{eq:mip_opt} contain $|\mathcal{P}|$ binary variables and $N+2$ structural constraints. After the three reductions, they contain $|\widetilde{\mathcal{P}}|$ binary variables and $G+2$ structural constraints.

\section{Experiments}
This section reports computational experiments evaluating the effectiveness of our method for stratified sampling.

\subsection{Experimental Setup}

We compared the performance of the following methods for variance reduction\footnote{Our implementation is available at \url{https://github.com/t087t/optimal-multiway-stratification-trees}.}:
\begin{itemize}
\item \textbf{CUPED}: A control-variate method using the covariate most highly correlated with the outcome~\cite{deng2013improving};
\item \textbf{K-means}: Stratified sampling based on $K$-means clustering with all candidate variables~\cite{kim2013stratified};
\item \textbf{K-means-DT}: CART trained to reproduce the $K$-means cluster labels~\cite{kim2013stratified};
\item \textbf{SFS-KM-V}: Stratified sampling via $K$-means clustering with variables sequentially selected to reduce the variance of the stratified estimator~\cite{momozu2025subset};
\item \textbf{CART}: A regression tree using mean-squared-error splits;
\item \textbf{OMST(quant)} and \textbf{OMST(optb)}: Our OMST methods with quantile binning and optimal binning, respectively.
\end{itemize}

For the stratified sampling methods, two allocation rules were evaluated: {\em Proportional} denotes the proportional allocation in Eq.~\eqref{eq:prop_alloc}, and {\em Optimal} denotes the constrained optimal allocation based on Eq.~\eqref{eq:allocation_obj}\text{--}\eqref{eq:allocation_st}. All methods were trained on a training dataset and evaluated on a held-out test dataset under the same sampling protocol. 
This data split can be viewed as learning fixed stratification rules from historical data and applying them to a future experimental population, although it does not reproduce a live A/B test.
Non-integer proportional sample sizes were rounded using the largest-remainder method; optimal allocation used $l_h=1$ and $u_h=N_h$ for all $h\in[H]$.

The evaluation was based on a Monte Carlo simulation protocol, in which the sample selection and the subsequent estimation of the population mean were repeated $T = 10{,}000$ times for each method. Let $\hat{\mu}^{(t)}$ be the sample mean obtained in the $t$-th simulation run, and let $\mu$ be the true population mean of the test dataset. The root mean squared error (RMSE) of the sample mean was defined as
\begin{equation}
\mathrm{RMSE}
:=
\sqrt{\frac{1}{T}\sum_{t=1}^{T}(\hat{\mu}^{(t)}-\mu)^2}.
\end{equation}
The RMSE quantifies the deviation of the estimator from the true population mean, offering the advantage of directly evaluating the estimation error, including both bias and variance, on the original scale of the outcome variable.

Let $\mathrm{RMSE}_{\mathrm{method}}$ and $\mathrm{RMSE}_{\mathrm{SRS}}$ denote the RMSEs of the sample mean under a given method and under simple random sampling, respectively. The primary evaluation metric was the RMSE reduction rate (\%), which measures the extent to which each method reduces the estimation error relative to the SRS baseline:
\begin{equation}
\text{RMSE Reduction (\%)}
:=
\left(
1-\frac{\mathrm{RMSE}_{\mathrm{method}}}{\mathrm{RMSE}_{\mathrm{SRS}}
}\right)\times 100.
\end{equation}

The sample size was $n = 100$. K-means, K-means-DT, and SFS-KM-V used five clusters, while K-means-DT and CART were limited to five leaves with a minimum leaf size of $5\%$ of the training data. 
Unless otherwise stated, OMST used a maximum depth of three, at most $L=5$ selected strata for the main comparison, and a path-pruning threshold of $N_{\min}=5\%$ of the training data. 
Increasing $L$ enlarges the feasible set of partitions but also increases optimization cost and may reduce interpretability, while the role of $N_{\min}$ follows the computational--statistical trade-off described in Sec.~\ref{sec:reduction}. 
We set $\rho = 1$ because estimating the population mean requires every unit to belong to a selected stratum; using $\rho<1$ would require a separate rule for uncovered units. 
In the depth-comparison experiment (depths one to five), OMST used $L=20$, while a CART tree of depth $d$ has at most $2^d$ leaves.
Quantile binning adopted four bins. Optimal binning was implemented via the \texttt{optbinning} package\footnote{\url{https://gnpalencia.org/optbinning/}}~\cite{navas2022optimal} using CART pre-binning with at most $20$ pre-bins, a minimum pre-bin size of $5\%$, a maximum $p$-value constraint of $0.05$ for merging adjacent bins, and the top three features selected by the binning quality score. 
K-means and CART were implemented using the Python scikit-learn library. All binary optimization problems were solved using Gurobi 12.0.2 with a one-hour time limit. Experiments were run in Python 3.13.2 on a Windows 11 workstation equipped with an Intel Core i7-1165G7 CPU and eight logical processors.

We used the following two datasets (one real-world and one simulated) with distinct characteristics:
\begin{itemize}
\item \textbf{Bike-sharing dataset}\footnote{\url{https://archive.ics.uci.edu/dataset/560/seoul+bike+sharing+demand}}: A real-world dataset recording hourly public bicycle rentals in Seoul alongside weather and holiday information. The outcome variable is the hourly rental count. Its 9 numerical and 3 categorical covariates yielded 15 features after one-hot encoding; the data were randomly split into training and test sets of $4{,}380$ units each.
\item \textbf{Customer-purchase dataset}\footnote{\url{https://www.kaggle.com/datasets/sanyamgoyal401/customer-purchases-behaviour-dataset}}: A synthetic dataset simulated to mimic consumer behavior scenarios. The outcome variable is the purchase amount. Its 2 numerical and 5 categorical covariates yielded 17 features after one-hot encoding; the data were randomly split into training and test sets of $50{,}000$ units each.
\end{itemize}

\subsection{Results}

\begin{figure}[tb]
\centering
\begin{minipage}{.49\linewidth}
\centering
\includegraphics[width=\linewidth]{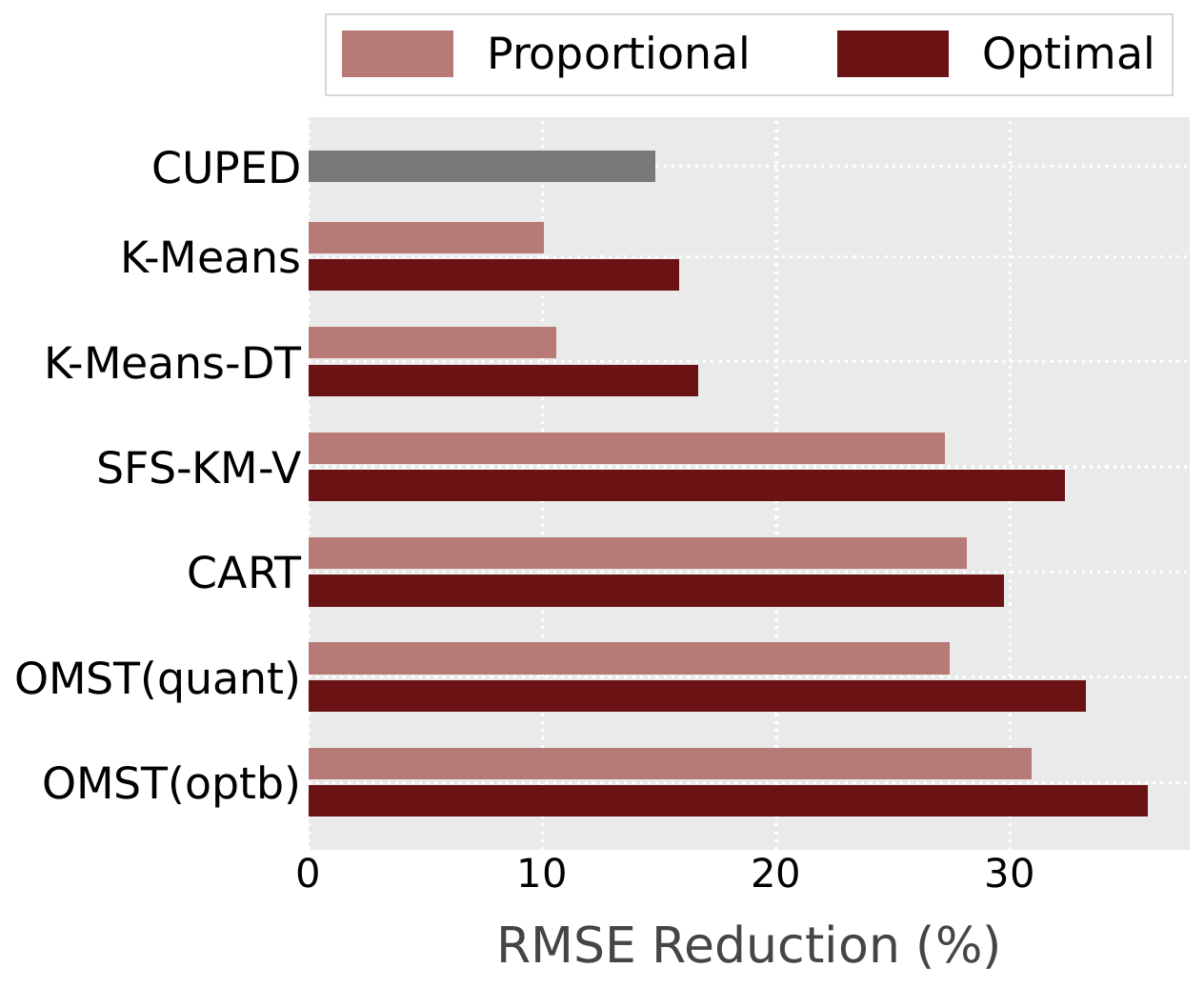}
{\small (a) Bike-sharing dataset}
\end{minipage}
\hfill
\begin{minipage}{.49\linewidth}
\centering
\includegraphics[width=\linewidth]{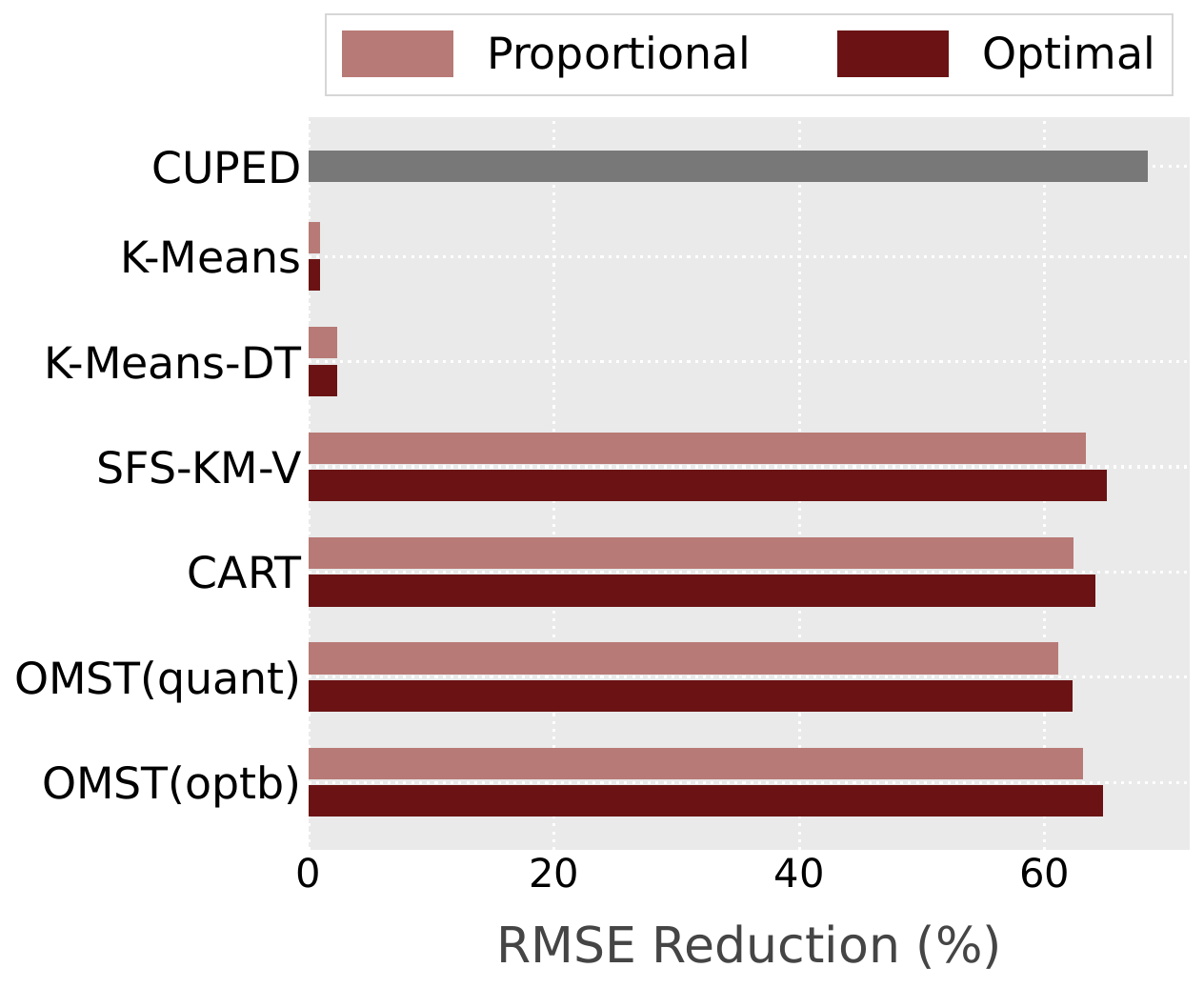}
{\small (b) Customer-purchase dataset}
\end{minipage}
\caption{RMSE reduction rates on the two datasets. Higher values indicate lower estimation error relative to SRS.}
\label{fig:rrr_comparison}
\end{figure}

Figure~\ref{fig:rrr_comparison} shows the RMSE reduction rates on the two datasets. On the bike-sharing dataset, OMST(optb) achieved the largest reduction among the compared stratification methods: $30.9\%$ under proportional allocation and $35.9\%$ under optimal allocation. On the customer-purchase dataset, CUPED achieved $68.4\%$, while OMST(optb) achieved $63.1\%$ and $64.8\%$, comparable to SFS-KM-V and CART. These results suggest that variance-based path selection is effective when multiple covariates jointly explain outcomes as in the bike-sharing dataset, whereas a single strong covariate can make regression adjustment highly competitive as in the customer-purchase dataset.

\begin{figure}[tb]
\centering
\begin{minipage}{.49\linewidth}
\centering
\includegraphics[width=\linewidth]{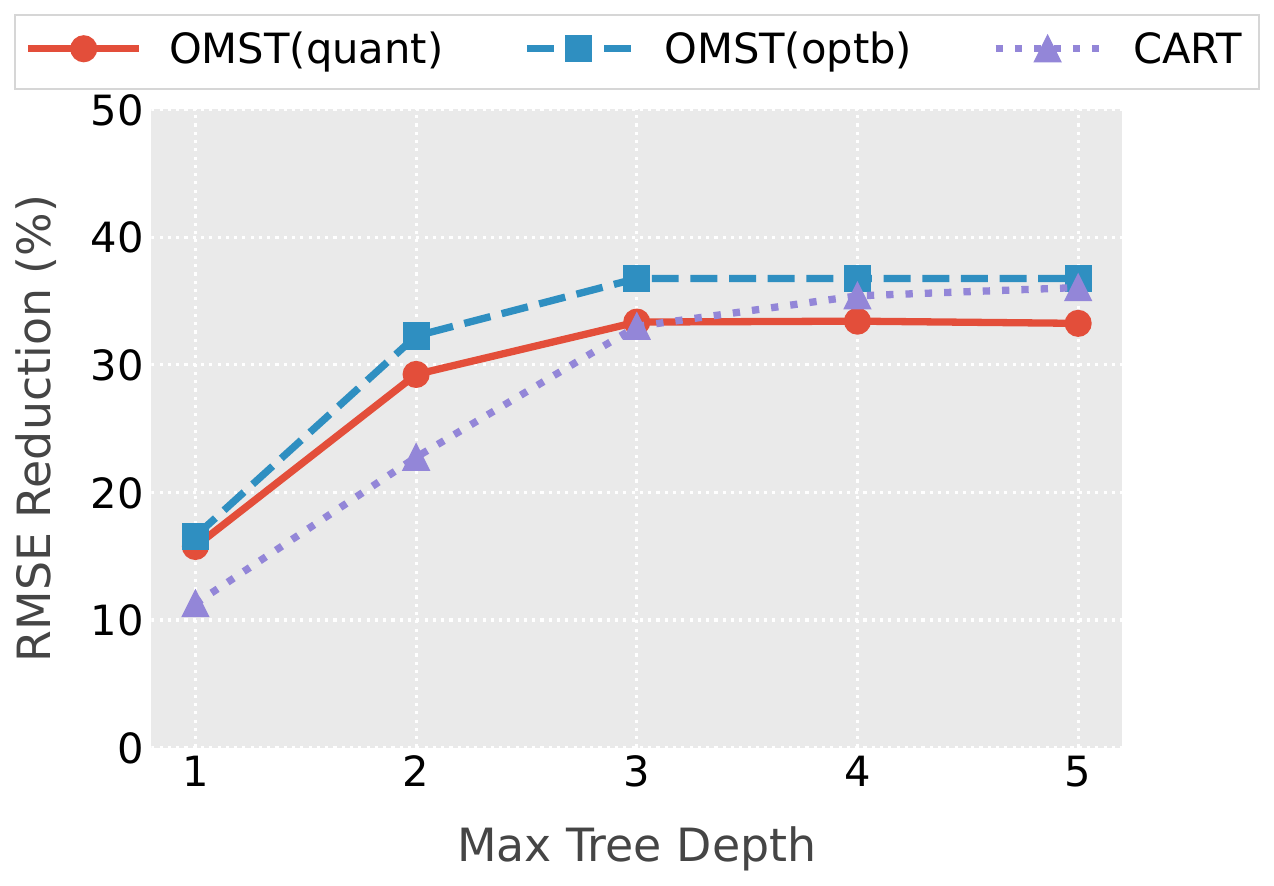}
{\small (a) Bike-sharing / Proportional}
\end{minipage}
\hfill
\begin{minipage}{.49\linewidth}
\centering
\includegraphics[width=\linewidth]{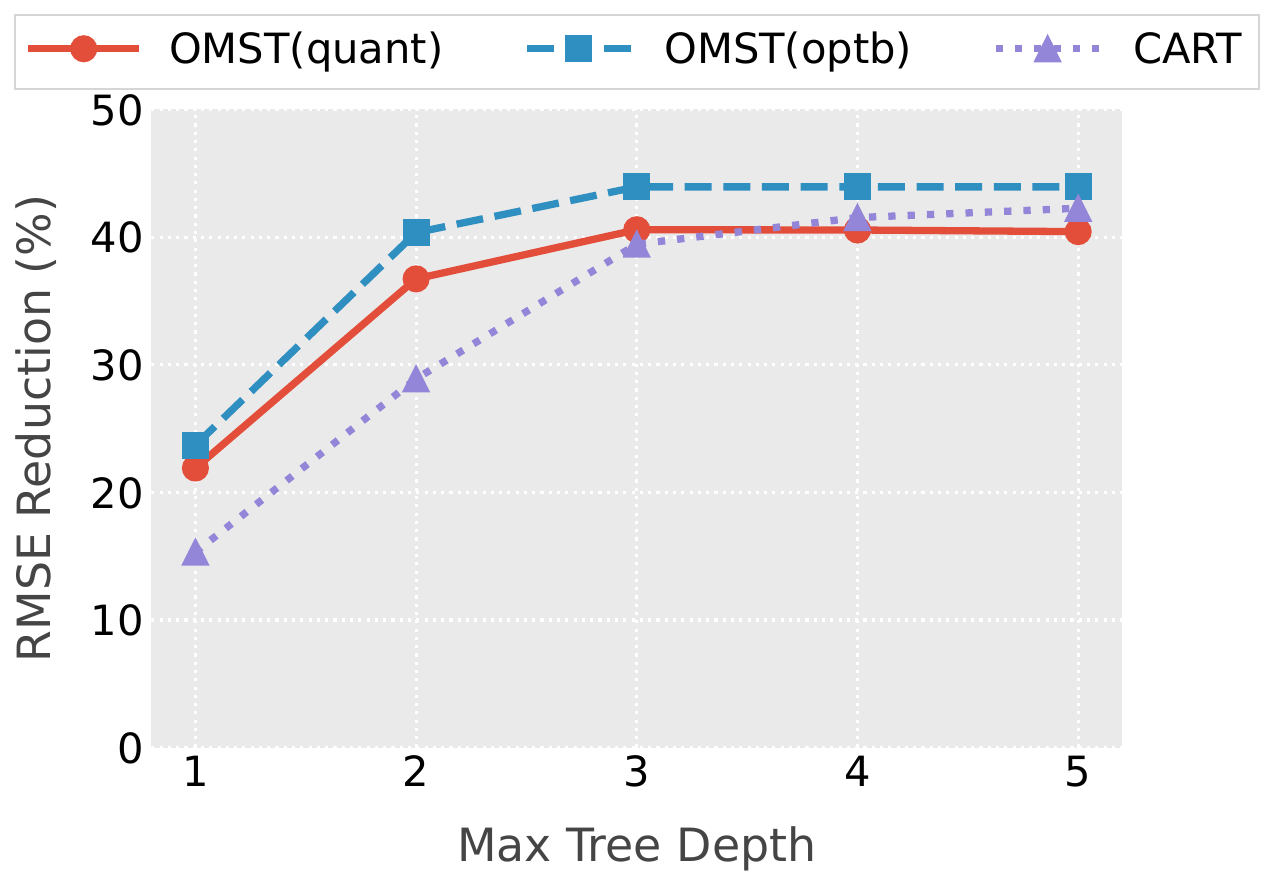}
{\small (b) Bike-sharing / Optimal}
\end{minipage}

\vspace{4mm}
\begin{minipage}{.49\linewidth}
\centering
\includegraphics[width=\linewidth]{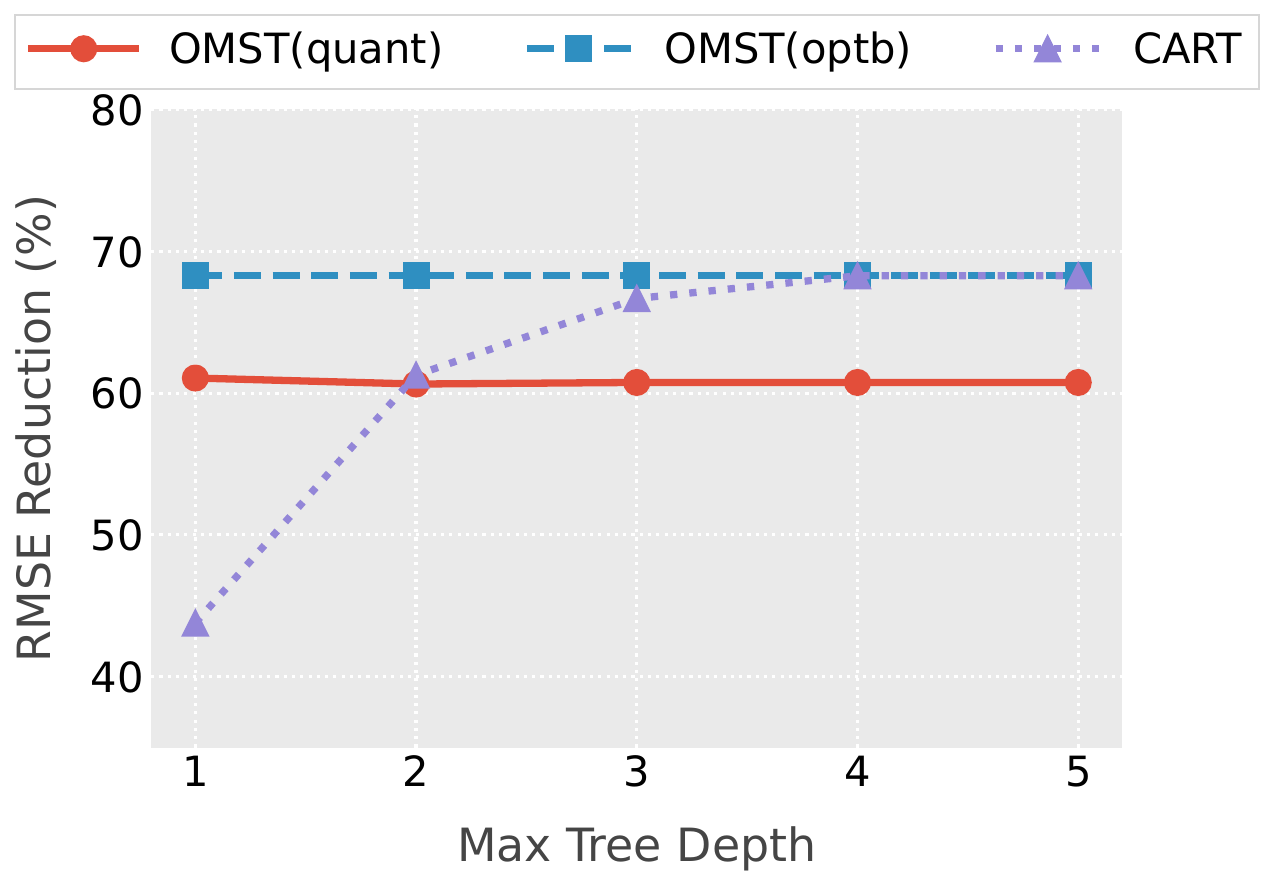}
{\small (c) Customer-purchase / Proportional}
\end{minipage}
\hfill
\begin{minipage}{.49\linewidth}
\centering
\includegraphics[width=\linewidth]{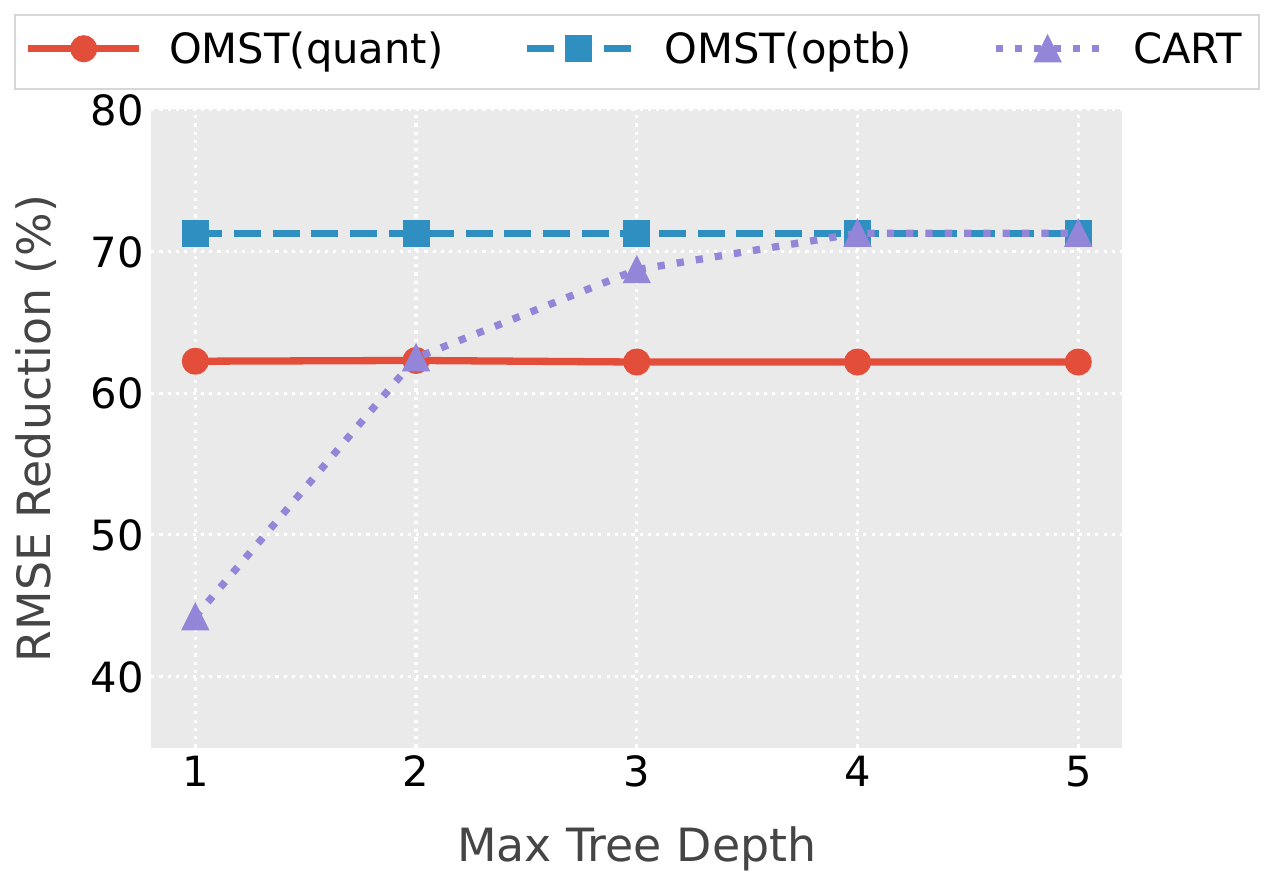}
{\small (d) Customer-purchase / Optimal}
\end{minipage}
\caption{RMSE reduction rates by maximum tree depth}
\label{fig:depth_comparison}
\end{figure}

Figure~\ref{fig:depth_comparison} highlights that OMST achieved strong RMSE reduction with shallow multi-way decision trees, which is central to its interpretability advantage over greedy binary-tree baselines. On the bike-sharing dataset, OMST(optb) showed increasing RMSE reduction up to depth three, reducing RMSE by $36.8\%$ under proportional and $43.9\%$ under optimal allocation. This saturation occurred because candidate path generation was limited to the top three features according to their binning quality scores, leaving no additional feature combinations beyond depth three. On the customer-purchase dataset, OMST(optb) achieved a large reduction at depth one, with deeper trees providing little further gain. Thus, while outcome-aware discretization enriches candidate paths, the effective depth depends on the structure of the data and the number of selected features.

Figure~\ref{fig:selected_trees} shows the stratification trees selected by OMST(optb) under optimal allocation. In Fig.~\ref{fig:selected_trees}(a), humidity defines the first split and temperature and hour refine one branch; each leaf is a sampling stratum rather than a prediction class. In Fig.~\ref{fig:selected_trees}(b), a shallow multi-way split on annual income suggests that this feature is particularly informative for within-stratum variance. This relationship is consistent with CUPED's strong RMSE reduction in Fig.~\ref{fig:rrr_comparison}, because CUPED is effective when one covariate strongly explains outcome variation.

\begin{figure}[tb]
\centering
\begin{minipage}{\linewidth}
\centering
\resizebox{0.95\linewidth}{!}{%
\begin{tikzpicture}[
  node distance=4mm and 4.5mm,
  every node/.style={draw, rounded corners=2pt, align=center, inner xsep=3pt,inner ysep=2.5pt,line width=.4pt, font=\scriptsize},
  edge/.style={-{Stealth[length=2mm]}, line width=.4pt},
  rootstyle/.style={fill=gray!30, minimum width=20mm},
  condhum/.style={fill=orange!15,  minimum height=6mm},
  condtemp/.style={fill=blue!15,  minimum height=6mm},
  condhour/.style={fill=green!15, minimum height=6mm},
  leaf/.style={fill=gray!5, minimum width=18mm}
]
\node[rootstyle] (root) {Source};

\node[condhum, below left=of root, xshift=-1mm] (h1) {Humidity$\le 88.500$};
\node[condhum, below right=of root, xshift=-8mm] (h2) {Humidity$> 88.500$};

\node[condtemp, below left=of h1, xshift=-1mm] (t1) {Temperature$> 7.850$};
\node[condtemp, below right=of h1, xshift=-10.5mm] (t2) {Temperature$\le 7.850$};

\node[condhour, below left=of t1, xshift=-1mm] (hr1) {Hour$\le 6.500$};
\node[condhour, below=of t1] (hr2) {$6.500 < \mathrm{Hour} \le 16.500$};
\node[condhour, below right=of t1, xshift=1mm] (hr3) {Hour$> 16.500$};

\node[leaf, below=of hr1] (p1) {Stratum 1\\$N=627$\\$s=289.280$};
\node[leaf, below=of hr2] (p2) {Stratum 2\\$N=1126$\\$s=471.016$};
\node[leaf, below=of hr3] (p3) {Stratum 3\\$N=811$\\$s=699.096$};
\node[leaf, below=of t2] (p4) {Stratum 4\\$N=1473$\\$s=265.664$};
\node[leaf, below=of h2] (p5) {Stratum 5\\$N=343$\\$s=269.622$};

\draw[edge] (root) -- (h1);
\draw[edge] (root) -- (h2);

\draw[edge] (h1) -- (t1);
\draw[edge] (h1) -- (t2);

\draw[edge] (t1) -- (hr1);
\draw[edge] (t1) -- (hr2);
\draw[edge] (t1) -- (hr3);

\draw[edge] (hr1) -- (p1);
\draw[edge] (hr2) -- (p2);
\draw[edge] (hr3) -- (p3);
\draw[edge] (t2) -- (p4);
\draw[edge] (h2) -- (p5);

\end{tikzpicture}
}

{\small (a) Bike-sharing dataset}
\end{minipage}

\vspace{4mm}

\begin{minipage}{\linewidth}
\centering
\resizebox{0.95\linewidth}{!}{%
\begin{tikzpicture}[
  node distance=5mm and 5.5mm,
  every node/.style={draw, rounded corners=2pt, align=center, inner xsep=3pt,inner ysep=2.5pt,line width=.4pt, font=\scriptsize},
  edge/.style={-{Stealth[length=2mm]}, line width=.4pt},
  rootstyle/.style={fill=gray!30, minimum width=20mm},
  condgreen/.style={fill=orange!15, minimum height=10mm},
  leaf/.style={fill=gray!5, minimum width=18mm}
]
\node[rootstyle] (root) {Source};

\node[condgreen, below=of root] (b3) {$21881.0 < $\\$\mathrm{Annual~income}$\\$\le 30854.5$};
\node[condgreen, left=of b3] (b2) {$13267.5 < $\\$\mathrm{Annual~income}$\\$\le 21881.0$};
\node[condgreen, left=of b2] (b1) {$\mathrm{Annual~income} $\\$\le 13267.5$};
\node[condgreen, right=of b3] (b4) {$30854.5 < $\\$\mathrm{Annual~income}$\\$\le 39246.5$};
\node[condgreen, right=of b4] (b5) {$\mathrm{Annual~income}$\\$> 39246.5$};

\node[leaf, below=of b1] (p1) {Stratum 1\\$N=9217$\\$s=960.685$};
\node[leaf, below=of b2] (p2) {Stratum 2\\$N=9642$\\$s=1252.346$};
\node[leaf, below=of b3] (p3) {Stratum 3\\$N=9741$\\$s=1601.005$};
\node[leaf, below=of b4] (p4) {Stratum 4\\$N=9437$\\$s=1928.828$};
\node[leaf, below=of b5] (p5) {Stratum 5\\$N=11963$\\$s=2499.509$};

\foreach \b in {b1,b2,b3,b4,b5} {
  \draw[edge] (root) -- (\b.north);
}

\foreach \b/\p in {b1/p1,b2/p2,b3/p3,b4/p4,b5/p5} {
  \draw[edge] (\b.south) -- (\p.north);
}

\end{tikzpicture}
}

{\small (b) Customer-purchase dataset}
\end{minipage}
\caption{Selected OMST(optb) stratification trees under optimal allocation. Each leaf denotes a selected stratum with stratum size $N$ and standard deviation $s$.}
\label{fig:selected_trees}
\end{figure}

Table~\ref{tab:speed_reduction} reports the combined impact of the problem reductions. Unit aggregation alone reduced the number of constraints from $N=4{,}380$ to $G=483$ for the bike-sharing dataset and from $N=50{,}000$ to $G=32$ for the customer-purchase dataset with optimal binning. For example, on the bike-sharing dataset, path pruning and aggregation jointly reduced the number of depth-three paths from $16{,}426$ to $14{,}970$ with quantile binning. Together, all reductions lowered the corresponding runtime from $69.3$ to $29.0$ seconds under proportional allocation and from $67.3$ to $33.5$ seconds under optimal allocation on the bike-sharing dataset.

\begin{table}[tb]
\centering
\caption{Impact of problem-size reduction on candidate paths, assignment patterns, and computation time}
\label{tab:speed_reduction}
\footnotesize
\setlength{\tabcolsep}{3.5pt}
\renewcommand{\arraystretch}{0.90}

{\centering
{\small (a) Bike-sharing dataset (\#Units$(N)=4,380$)} \\[1mm]
\begin{tabular}{cc r rr r r rr}
\toprule
 & & \multicolumn{3}{c}{Before Reduction} & \multicolumn{4}{c}{After Reduction} \\
\cmidrule(lr){3-5} \cmidrule(lr){6-9}
 & & & \multicolumn{2}{c}{Time (s)} & & & \multicolumn{2}{c}{Time (s)} \\
\cmidrule(lr){4-5} \cmidrule(lr){8-9}
Binning & Depth & \#Paths & Prop. & Opt. & \#Units($G$) & \#Paths & Prop. & Opt. \\
\midrule
Quantile & 1 & 62     & 0.1  & 0.2  & 1,837 & 62     & $<0.1$ & 0.1 \\
 & 2 & 1,578  & 9.3  & 1.4  & 1,837 & 1,541  & 4.9    & 1.1 \\
 & 3 & 16,426 & 69.3 & 67.3 & 1,837 & 14,970 & 29.0   & 33.5 \\
\cmidrule(lr){1-9}
Optimal & 1  & 127    & $<0.1$ & 0.1  & 483   & 127    & $<0.1$ & $<0.1$ \\
 &  2 & 3,670  & 3.1  & 2.3  & 483   & 3,667  & 2.0    & 1.1 \\
 &  3 & 27,497 & 15.5 & 20.9 & 483   & 27,439 & 9.1    & 13.5 \\
\bottomrule
\end{tabular}
\par}

\vspace{5mm}

{\centering
{\small (b) Customer-purchase dataset (\#Units$(N)=50,000$)} \\[1mm]
\begin{tabular}{cc r rr r r rr}
\toprule
 & & \multicolumn{3}{c}{Before Reduction} & \multicolumn{4}{c}{After Reduction} \\
\cmidrule(lr){3-5} \cmidrule(lr){6-9}
 & & & \multicolumn{2}{c}{Time (s)} & & & \multicolumn{2}{c}{Time (s)} \\
\cmidrule(lr){4-5} \cmidrule(lr){8-9}
Binning & Depth & \#Paths & Prop. & Opt. & \#Units($G$) & \#Paths & Prop. & Opt. \\
\midrule
Quantile & 1 & 34    & 0.6 & 0.5 & 4,373 & 34    & $<0.1$ & 0.1 \\
 & 2 & 438   & 3.2 & 2.5 & 4,373 & 438   & 1.9    & 0.4 \\
 & 3 & 1,637 & 19.0 & 5.9 & 4,373 & 1,637 & 5.4    & 1.4 \\
\cmidrule(lr){1-9}
Optimal &  1 & 137   & 1.5 & 1.7 & 32    & 137   & $<0.1$ & $<0.1$ \\
 &  2 & 362   & 2.6 & 3.0 & 32    & 362   & $<0.1$ & $<0.1$ \\
 &  3 & 362   & 2.7 & 2.9 & 32    & 362   & $<0.1$ & $<0.1$ \\
\bottomrule
\end{tabular}
\par}
\end{table}

\section{Conclusion}
We proposed Optimal Multi-way Stratification Trees (OMST), an optimization-based stratification method for online controlled experiments. Our method selects interpretable directed paths using an exact variance-minimizing objective under continuous proportional allocation and a Neyman-type optimal allocation. Supervised optimal binning generates outcome-relevant candidate paths, while problem reduction procedures for redundant paths and equivalent assignment patterns improve computational scalability.

Experiments on the bike-sharing and customer-purchase datasets showed that OMST can achieve comparable or superior RMSE reduction while maintaining shallow, interpretable trees. OMST with optimal binning achieved the best reduction among the stratification methods on the bike-sharing dataset. On the customer-purchase dataset, it performed comparably to SFS-KM-V and CART, while CUPED was highly effective because a single covariate was strongly correlated with the outcome. The selected OMST trees were more directly interpretable than CART-based binary trees because multi-way splits avoided deeply nested conditions. The benefit of greater depth was dataset dependent, and the unit aggregation substantially reduced the number of constraints.

This study is limited to one real-world and one synthetic dataset, and held-out resampling does not fully reproduce a live A/B testing system. In practice, OMST can be trained offline and its shallow rules applied online at relatively low assignment cost. However, distribution shifts require drift monitoring and periodic retraining. Moreover, independent feature-wise discretization may miss interaction-specific thresholds, while exhaustive path enumeration limits high-dimensional scalability.

Future work includes broader real-world validation, comparison with modern machine-learning-based regression adjustment methods, sensitivity analyses of model parameters, and a detailed ablation of the reduction procedures. Further directions include joint feature discretization and dynamic path generation, such as column generation.

\FloatBarrier

\bibliographystyle{splncs04}
\bibliography{mybibliography}

@article{costa2023recent,
  title   = {Recent advances in decision trees: An updated survey},
  author  = {Costa, Vinicius G. and Pedreira, Carlos E.},
  journal = {Artificial Intelligence Review},
  volume  = {56},
  number  = {5},
  pages   = {4765--4800},
  year    = {2023},
  publisher = {Springer}
}

@inproceedings{momozu2025subset,
  author    = {Momozu, Haru and Uehara, Yuki and Nishimura, Naoki and Ohashi, Koya and Jobson, Deddy and Li, Yilin and Dinh, Phuong and Sukegawa, Noriyoshi and Takano, Yuichi},
  title     = {Subset Selection for Stratified Sampling in Online Controlled Experiments},
  booktitle = {PRICAI 2025: Trends in Artificial Intelligence},
  series    = {Lecture Notes in Computer Science},
  volume    = {16452},
  pages     = {600--613},
  year      = {2026},
  publisher = {Springer}
}

@article{suzuki2026interpretable,
  author  = {Suzuki, H. and Ikeda, S. and Nishimura, N. and Takano, Y.},
  title   = {Interpretable clustering via optimal multi-way decision trees},
  journal = {arXiv preprint arXiv:2602.13586},
  year    = {2026}
}

@inproceedings{xie2016improving,
  title={Improving the Sensitivity of Online Controlled Experiments: Case Studies at {Netflix}},
  author={Xie, Huizhi and Aurisset, Juliette},
  booktitle={Proceedings of the 22nd ACM SIGKDD International Conference on Knowledge Discovery and Data Mining},
  pages={645--654},
  year={2016},
  publisher={ACM}
}

@article{friedrich2015fast,
  title={Fast integer-valued algorithms for optimal allocations under constraints in stratified sampling},
  author={Friedrich, Ulf and M{\"u}nnich, Ralf and de Vries, Sven and Wagner, Matthias},
  journal={Computational Statistics \& Data Analysis},
  volume={92},
  pages={1--12},
  year={2015}
}

@article{tabord2023stratification,
  title={Stratification Trees for Adaptive Randomisation in Randomised Controlled Trials},
  author={Tabord-Meehan, Max},
  journal={The Review of Economic Studies},
  volume={90},
  number={5},
  pages={2646--2673},
  year={2023}
}

@article{kim2013stratified,
  title={Stratified sampling design based on data mining},
  author={Kim, Yong Joon and Oh, Young Jae and Park, Seung Ho and Cho, Sungzoon and Park, Hyunsoo},
  journal={Healthcare Informatics Research},
  volume={19},
  number={3},
  pages={186--195},
  year={2013}
}

@inproceedings{subramanian2023scalable,
  title={Scalable Optimal Multiway-Split Decision Trees with Constraints},
  author={Subramanian, Shivaram and Sun, Wei},
  booktitle={Proceedings of the AAAI Conference on Artificial Intelligence},
  volume={37},
  pages={9891--9899},
  year={2023}
}

@article{navas2022optimal,
  title={Optimal binning: Mathematical programming formulation},
  author={Navas-Palencia, Guillermo},
  journal={arXiv preprint arXiv:2001.08025},
  year={2022}
}

@inproceedings{deng2013improving,
  title={Improving the sensitivity of online controlled experiments by utilizing pre-experiment data},
  author={Deng, Alex and Xu, Ya and Kohavi, Ron and Walker, Toby},
  booktitle={Proceedings of the Sixth ACM International Conference on Web Search and Data Mining},
  pages={123--132},
  year={2013}
}

@article{steinley2006k,
  title={K-means clustering: A half-century synthesis},
  author={Steinley, Douglas},
  journal={British Journal of Mathematical and Statistical Psychology},
  volume={59},
  number={1},
  pages={1--34},
  year={2006}
}

@inproceedings{guo2021machine,
  title={Machine Learning for Variance Reduction in Online Experiments},
  author={Guo, Yongyi and Coey, Dominic and Konutgan, Mikael and Li, Wenting and Schoener, Chris and Goldman, Matt},
  booktitle={Advances in Neural Information Processing Systems},
  volume = {34},
  pages={8637--8648},
  year={2021}
}

@article{freedman2008regression,
  title={On regression adjustments in experiments with several treatments},
  author={Freedman, David A},
  journal={The Annals of Applied Statistics},
  volume={2},
  number={1},
  pages={176--196},
  year={2008}
}

@article{holt1979post,
  title={Post Stratification},
  author={Holt, D. and Smith, T. M. F.},
  journal={Journal of the Royal Statistical Society. Series A (General)},
  volume={142},
  number={1},
  pages={33--46},
  year={1979}
}

@book{Cochran1977,
  title={Sampling Techniques},
  author={Cochran, William G.},
  edition={3},
  year={1977},
  publisher={John Wiley \& Sons},
  address={New York}
}

@book{breiman1984cart,
  title     = {Classification and Regression Trees},
  author    = {Breiman, Leo and Friedman, Jerome H. and Olshen, Richard A. and Stone, Charles J.},
  year      = {1984},
  publisher = {Wadsworth International Group},
  address   = {Belmont, CA}
}

@article{Neyman1934,
  author  = {Neyman, Jerzy},
  title   = {On the two different aspects of the representative method: The method of stratified sampling and the method of purposive selection},
  journal = {Journal of the Royal Statistical Society},
  volume  = {97},
  number  = {4},
  pages   = {558--625},
  year    = {1934}
}

@article{Tipton2014,
  title   = {Stratified Sampling Using Cluster Analysis: A Sample Selection Strategy for Improved Generalizations From Experiments},
  author  = {Tipton, Elizabeth},
  journal = {Evaluation Review},
  volume  = {37},
  number  = {2},
  pages   = {109--139},
  year    = {2014}
}

@article{charette2025improving,
  title  = {Improving Sensitivity in {A/B} Tests: Integrating {CUPED} with Trimmed Mean Techniques},
  author = {Charette, Kevin and Boudreault, Tristan},
  journal = {arXiv preprint arXiv:2510.03468},
  year   = {2025}
}

@article{bertsimas2017optimal,
  title   = {Optimal Classification Trees},
  author  = {Bertsimas, Dimitris and Dunn, Jack},
  journal = {Machine Learning},
  volume  = {106},
  number  = {7},
  pages   = {1039--1082},
  year    = {2017},
  publisher = {Springer}
}

@inproceedings{aghai2021flowoct,
  title     = {Flow-Based Optimal Classification Trees},
  author    = {Aghaei, Saeed and Gomez, Andres and Vayanos, Phebe},
  booktitle = {Proceedings of the AAAI Conference on Artificial Intelligence},
  volume    = {35},
  pages     = {5686--5694},
  year      = {2021}
}

@inproceedings{dougherty1995supervised,
  title={Supervised and unsupervised discretization of continuous features},
  author={Dougherty, James and Kohavi, Ron and Sahami, Mehran},
  booktitle={Proceedings of the Twelfth International Conference on Machine Learning},
  pages={194--202},
  year={1995}
}

\end{document}